\documentclass[12pt]{article}
\usepackage[T1]{fontenc}
\usepackage{arabtex}

\usepackage{coling2018}

\usepackage{url}
\usepackage[para,online,flushleft]{threeparttable}
\usepackage{booktabs}
\usepackage{float}
\usepackage{hhline}
\usepackage{graphicx,subcaption}
\usepackage{smartdiagram}
\usepackage{pgf-pie}
\usepackage{multicol}
\usepackage{enumitem}
\usepackage{caption} 

\usepackage{colortbl}

\usepackage[stable]{footmisc}

\usepackage{cp1256,utf8}
\setcode{utf8}

\date{}

\definecolor{lightgray}{rgb}{0.95,0.95,0.95}

\begin{document}

\title{Mining Mental Health Signals: A Comparative Study of Four Machine Learning Methods for Depression Detection from Social Media Posts in Sorani Kurdish}

\author{
	\begin{tabular}[t]{c}
		Idrees Mohammed and Hossein Hassani\\
		\textnormal{University of Kurdistan Hewl\^er}\\
		\textnormal{Kurdistan Region - Iraq}\\
		{\tt {\{idrees.mohammed, hosseinh}\}@ukh.edu.krd}
	\end{tabular}
}

\maketitle

\begin{abstract}
Depression is a common mental health condition that can lead to hopelessness, loss of interest, self-harm, and even suicide. Early detection is challenging due to individuals not self-reporting or seeking timely clinical help. With the rise of social media, users increasingly express emotions online, offering new opportunities for detection through text analysis. While prior research has focused on languages such as English, no studies exist for Sorani Kurdish. This work presents a machine learning and Natural Language Processing (NLP) approach to detect depression in Sorani tweets. A set of depression-related keywords was developed with expert input to collect 960 public tweets from X (Twitter platform)\footnote{We use X and Twitter interchangeably to refer to the same platform throughout this paper.}. The dataset was annotated into three classes: Shows depression, Not-show depression, and Suspicious by academics and final year medical students at the University of Kurdistan Hewlêr. Four supervised models, including Support Vector Machines, Multinomial Naive Bayes, Logistic Regression, and Random Forest, were trained and evaluated, with Random Forest achieving the highest performance accuracy and F1-score of 80\%. This study establishes a baseline for automated depression detection in Kurdish language contexts.
\end{abstract}

\section{Introduction}
Depression is a mental health condition that affects the daily life of individuals globally, and it causes individuals to lose interest in things that used to be interesting to them, feel low, have difficulty in making decisions, self-harm, attempt suicide~\cite{effectsofdepression}, academic performance~\cite{BlackDeer}, and academic dropout~\cite{SINVAL2025665}. Early detection is critical to improving quality of life and preventing severe outcomes, yet conventional approaches often rely on self-reporting or clinical assessments, which delay intervention. With the growth of social media, user-generated text offers an alternative channel for identifying depressive symptoms through computational analysis \cite{ReviewOnDepression,Tejaswini}.

While machine learning (ML) and natural language processing (NLP) have been applied to depression detection in languages such as English~\cite{JOSHI2022217,lorenzoni2025assessing}, Sorani Kurdish (Central Kurdish) remains underexplored in this area. Sorani is spoken by several million people in Iraq and neighboring Kurdish-speaking regions, but suffers from limited NLP resources, the scarcity of annotated datasets, and challenges such as mixed Persian-Arabic and Latin scripts \cite{SentementAnalysisKurdish}.

This research addresses these gaps by creating a Sorani depression dataset from social media posts and developing supervised ML models, including Support Vector Machines (SVM), Logistic Regression (LR), Multinomial Naive Bayes (NB), and Random Forest (RF), for automated depression detection. The main objectives are:
\begin{enumerate}
    \item Build and annotate a dataset of Sorani posts indicating depressive symptoms.
    \item Train and evaluate multiple ML models for classification.
    \item Assess model performance to establish a baseline for future research.
\end{enumerate}

The scope of this study is limited to social media text in Sorani, focusing on linguistic patterns that may indicate depression. Constraints include data scarcity, limited language tools, and ethical considerations in handling user-generated content.

The rest of this paper is organized as follows. Section 2 provides the related work, Section 3 describes the research method, Section 4 presents the results, and Section 5 concludes the research and provides future expansion areas.

\section{Related Work}
Most early work on depression detection has been conducted for texts in English. For example, \newcite{DeChoudhury} used 2.1M tweets with an SVM classifier, achieving 70\% accuracy, while \newcite{NadeemDep} trained Naive Bayes, SVM, Logistic Regression, and other models on 2.5M tweets, with unigram Naive Bayes reaching 86\% accuracy. Similarly, \newcite{AldarweshLR} applied Random Forest to LiveJournal posts, achieving 90\% accuracy. Deep learning approaches have also emerged, for instance, \newcite{DeepLearning} combined Long Short-Term Memory (LSTM) and Recurrent Neural Network (RNN) to achieve 99\% accuracy on an English Twitter dataset. Extending beyond depression, \newcite{KadkhodaLR} explored bipolar disorder detection using Twitter data, leveraging machine learning and NLP techniques to analyze user content and identify patterns. They employed various classifiers, including SVM, RF, NB, LR, Decision Tree (DT), K-Nearest Neighbor (KNN), and Stochastic Gradient Descent (SGD), with RF and NB both achieving 86\% accuracy. Their study also highlighted the potential of social media for mental health monitoring, while noting risks related to misuse, bias, and fairness in online data–driven assessments.

Research in other languages shows similar trends. Persian language studies have used Convolutional Neural Network (CNN), LSTM, and transformer models, with XLM-Roberta-large achieving an F1 score of 75.39\% \cite{MirzaLR}. Arabic studies report accuracies between 82\%–88\% using models such as Random Forest, KNN, and Attention Bi-LSTM \cite{AlaskarLR,AlmarsLR,musleh2022depressionLR}. Roman Urdu-English bilingual datasets have also been explored, with SVM achieving 84\% accuracy \cite{rehmaniLR}. Overall, traditional ML methods such as SVM, NB, and RF remain competitive for small to medium datasets, while deep learning and transformer-based models dominate in larger, more balanced datasets.

Kurdish remains a low-resource language in NLP, with no prior work on depression detection from text. Related efforts include sentiment analysis datasets for Sorani and Kurmanji dialects, such as works by \newcite{SamiLR}, \newcite{MahmudLR}, and \newcite{BadawiLR}, achieving accuracies between 78\%–89\% using SVM, Naive Bayes, and XLM-R. These works highlight challenges such as limited resources, dialectal variation, and mixed scripts, which complicate preprocessing and model training.

Across languages, X (formerly known as Twitter) is the most common data source for depression related research. Dataset sizes vary widely from a few thousand posts in low-resource languages to millions in English affecting model choice and performance. Classical ML approaches perform well with smaller datasets, while deep learning and transformer based architectures yield superior results when sufficient data and computational resources are available. The absence of Sorani depression detection studies underscores the novelty and significance of the present work.

\section{Methodology}
This study applies machine learning (ML) and natural language processing (NLP) to detect depression from Sorani social media posts. The methodology consists of data collection, annotation, preprocessing, feature extraction, model training, and evaluation \par
\begin{figure}[ht!]
		\centering
		\fbox{	\includegraphics[scale=0.8]{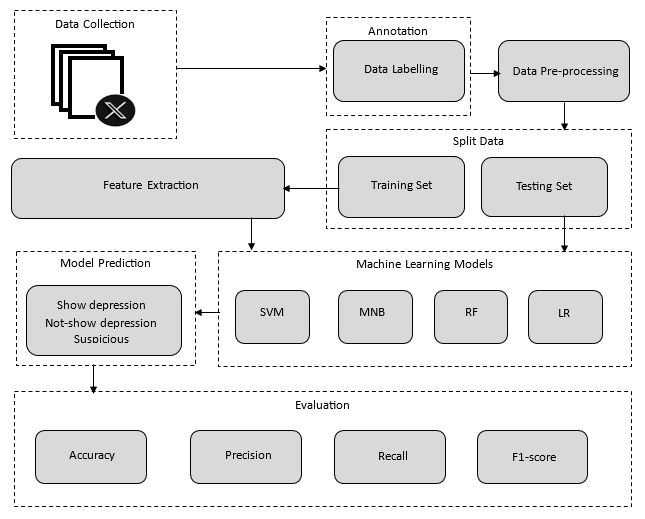}}
		\caption{Research methodology }
		\label{fig:ResearchMethodology}
		
	\end{figure}

 Figure \ref{fig:ResearchMethodology} present the flow of the methodology.

\subsection{Data Collection}
We consult psychiatrists and mental health professionals to identify key depressive symptoms, refine keyword selection, and validate the relevance of the data collection approach.

Due to the absence of existing Sorani depression datasets, we will collect data from X, a widely used platform for short, emotion rich posts. A set of ten expert verified depression-related keywords as shown in table \ref{tab:Highlighted-keywords} guided the search process. Data retrieval going to be through using the Twitter (currently known as X) API and custom Python scripts, with personally identifiable information removed in compliance with ethical and platform guidelines.
    
    \begin{table}[ht!]
		\begin{center}
			\caption{Highlighted keywords used for searching content on the Twitter platform}
			\begin{tabular}{|c|r|}
				\hline
				\rowcolor{lightgray}
				\textbf{\#} & \textbf{Highlighted keywords} \\ \hline
				1 & \RL{دڵتەنگی}                \\ \hline
				2 & \RL{نەخۆشی دەرونی}                \\ \hline
				3 & \RL{بێهیوایی}                \\ \hline
				4 & \RL{خەمۆکی}                \\ \hline
				5 &\RL{خۆکوشتن}                \\ \hline
				6  & \RL{بێزاری لە ژیان}                \\ \hline
				7  & \RL{بێتاقەت}                \\ \hline
                8  & \RL{پەشیمانی }                \\ \hline
                9  & \RL{نائومێدی}                \\ \hline
                10  & \RL{غەمباری}                \\ \hline

			\end{tabular}
			\label{tab:Highlighted-keywords}
		\end{center}
	\end{table}

\subsection{Annotation}
The dataset will manually labeled by medical experts and final year medical students at the University of Kurdistan Hewlêr into three classes: Shows depression, Not-show depression, and Suspicious.

\subsection{Preprocessing}
Tweets will be normalized, anonymized, and cleaned to remove URLs, symbols, English words, emojis, numbers, and duplicates. The Kurdish Language Processing Toolkit (KLPT) is used for normalization and standardization, addressing script inconsistencies in Sorani.

\subsection{Feature Extraction}
We will transform texts into numerical vectors using Term Frequency-Inverse Document Frequency (TF-IDF), which assigns higher weights to rare but informative terms while down weighting frequent, less informative words.

\subsection{Data Splitting}
We will split the dataset into 90\% training and 10\% testing, with 10-fold cross-validation applied to improve robustness and reduce overfitting.

\subsection{Machine Learning Models}
We will trained and evaluated four supervised classifiers including:

Support Vector Machine (SVM): effective for high dimensional text classification.

Multinomial Naive Bayes (MNB): a probabilistic model well suited for word frequency features.

Random Forest (RF): an ensemble of decision trees to improve accuracy and reduce overfitting.

Logistic Regression (LR): a linear model producing class probabilities.

\subsection{Evaluation}
We will evaluate models using Accuracy, Precision, Recall, and F1-score, with confusion matrices for class-level performance. Cross-validation ensured consistency between training and testing performance.

\section{Results and Discussion}
This chapter presents the dataset description, pre-processing, visualization, experimental implementation, performance evaluation, and discussion of machine learning models for detecting depression from Sorani social media texts.

\subsection{Data Collection}
Due to initial challenges in engaging mental health experts unfamiliar with ML and NLP, a consultation was held with a psychiatry expert with a background in computer science. This session validated the study’s psychological foundations, refined depression-related keywords, and provided key references, significantly enhancing the study’s rigor.

Using the Twitter API and Python scripts, 960 Sorani tweets related to depression were collected, based on psychiatrist approved keywords (e.g., sadness, mental illness, suicide). Ethical standards were maintained by anonymizing data and respecting platform policies.

\subsection{Annotation Process}
Manual annotation was performed by senior medical students under expert supervision, labeling tweets into three classes: Show Depression, Not-show Depression, and Suspicious. The labeled dataset was consolidated for analysis.

\par
\begin{figure}[H]
    \centering    \fbox{\includegraphics[width=0.877\linewidth]{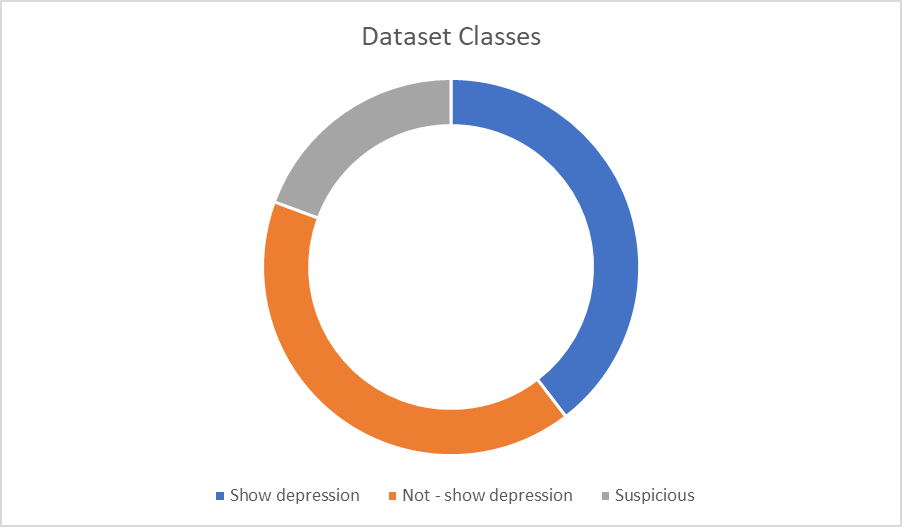}}
    \caption{Amount of data of each dataset class}
    \label{fig:Pie}
\end{figure}

 Figure \ref{fig:Pie} shows the visual representation of the amount of data of each class from the dataset.

\par  Table \ref{tab:Sample of the data for all classes of the dataset} Shows a sample of data for each class of the dataset.

\begin{table}[H]
    \centering
    \caption{Sample of the data for all classes of the dataset}
    \renewcommand{\arraystretch}{1.5} 
    \begin{tabular}{|p{4.7cm}|p{4.7cm}|p{4.7cm}|}
        \hline
        \rowcolor{lightgray}
        \textbf{Show depression} & \textbf{Not-show depression} & \textbf{Suspicious} \\ \hline
        
        \RL{من نەخۆشییەکی تەندروستی دەروونیم هەیە بە ناوی خەمۆکی لە بنەڕەتدا مانای ئەوەیە زۆر دڵتەنگم و سەختە باشتر بم} & 
        \RL{وەرزشکردن بۆ ماوەی تەنها خولەک لە ڕۆژێکدا ئەگەری تووشبوون بە خەمۆکی بۆ یەک لەسەر سێ کەمدەکاتەوە بەپێی دوایین لێکۆڵینەوەکان} & 
        \RL{چۆن زاڵ بین بەسەر دڵەڕاوکێی پەیوەست بە خەمۆکی لە کاتی هەوڵدان بۆ خوێندن} \\ \hline

        \RL{خراپترین شت لە خەمۆکیدا ئەوەیە کە هەمیشە هەست بە بەتاڵی یان تەنیایی دەکەیت بەبێ گوێدانە هەر شتێک
} &
        \RL{من زۆرم لۆرن خۆشدەوێت ئەو یارمەتی داوم لە خەمۆکی و دڵەڕاوکێ و سەختترین کاتەکاندا لەگەڵ خێزان و هاوڕێکانم زۆر سوپاس بۆ هەموو شتێک
} &
        \RL{پێشتر یاری تۆپی پێم دەکرد و دواتر لە پۆلی  تووشی خەمۆکی بووم
} \\ \hline

        \RL{هەمووان پێت دەڵێن واز لە خەمۆکی بێنە بەڵام ناتوانیت خەمۆکی واز لێبێنیت
} &
        \RL{سۆشیال میدیا خەڵکێکی وای تێدایە کە وا بیر دەکەنەوە خەمۆکی مرۆڤ دەکوژێت
} &
        \RL{بۆچی پێی دەڵێن ژەمێکی خۆش ئەگەر تامێکی تەواوی وەک خەمۆکی هەبێت
} \\ \hline

        \RL{خۆزگە خۆکوشتن حەڵاڵ ئەبو ئەوکاتە قەت چاوەڕیی گەورەبونی خەمەکانم نەئەکرد چونکە ئیتر توانای بەردەوام بونی ژیانم نەما} &
        
        \RL{مۆز میوەیەکی خۆشە خواردنی تەنها یەکێکیان دەتوانێت یارمەتیدەر بێت بۆ ڕزگاربوون لە هەست و سۆزی توڕەیی
} &
        \RL{من زۆر پەرۆش دەبم بۆ خەوتن ئەوە نیشانەیە بۆ خەمۆکی
} \\ \hline

        \RL{من بەدەست خەمۆکییەوە دەناڵێنم} &
        
        \RL{هەستێکی خۆشە کە ئەزیزانت لەلات بن و یارمەتیت بدەن و نەیەڵن تەنیایی یەخەت پێ بگرێ
} &
        \RL{شەوخوێیە بۆ سەر برینی دڵ تەنگی
} \\ \hline

    \end{tabular}
    \label{tab:Sample of the data for all classes of the dataset}
\end{table}

\par After completion of data labeling, the dataset is formed and ready for preprocessing and analysis.

\par
\begin{figure}[H]
    \centering    \fbox{\includegraphics[width=0.877\linewidth]{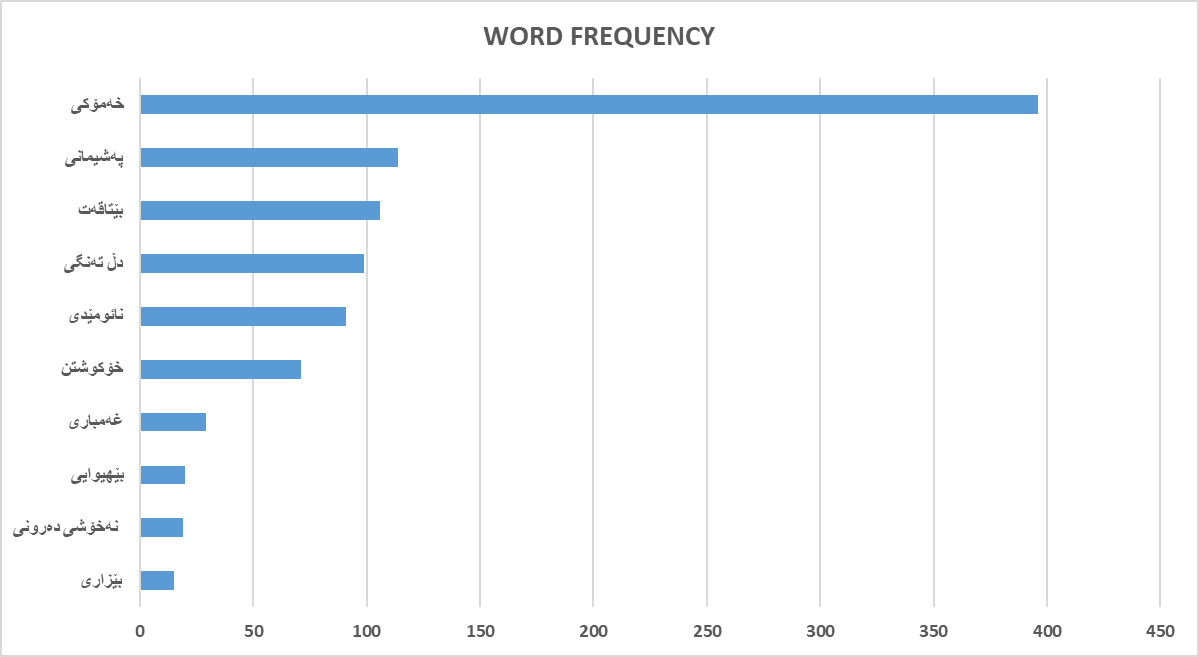}}
    \caption{Word frequency visualization of highlighted keywords from the dataset}
    \label{fig:Wordfrequency}
\end{figure}

 Figure \ref{fig:Wordfrequency} Shows the visual representation of the frequency of highlighted keywords from the dataset.

\subsection{Data Preprocessing}
Preprocessing included normalization and standardization (using KLPT~\cite{ahmadi-2020-klpt}), removal of English words, URLs, punctuation, emojis, numeric digits, extra spaces, and duplicate records, resulting in a clean dataset ready for modeling.

\subsection{Experimental Setup}
Experiments were implemented in Python 3.13.2 using Jupyter Notebook. Core libraries included Scikit learn for ML models, Pandas for data handling, Regular Expressions for text processing, and Seaborn/Matplotlib for visualization. Text was vectorized using TF-IDF to capture word importance.

\subsection{Dataset Splitting and Validation}
The dataset was split 90\% training and 10\% testing. To improve generalizability and reduce overfitting, 10-fold cross-validation was employed, ensuring that each data point was used for both training and validation across iterations.

\subsection{Results of Experiments}

Four experiments were conducted using Multinomial Naive Bayes (MNB), Support Vector Machine (SVM), Logistic Regression (LR), and Random Forest (RF) to evaluate classification performance on various dataset configurations.

\subsubsection{First Experiment: Binary Classification (Suspicious Class Removed)}
Removing the ‘Suspicious’ class reduced the dataset to 732 tweets (363 Show Depression, 369 Not-show Depression). SVM achieved the best performance with precision, recall, accuracy, and F1-score approximately 0.667. MNB followed closely, while LR had the lowest scores. Confusion matrices showed moderate misclassification between classes.

\subsubsection{Second Experiment: Imbalanced Dataset (All Classes)}
Using the full dataset of 911 tweets (including Suspicious), model performance declined. SVM had the highest recall (0.518) but a lower F1-score (0.476). MNB showed the highest precision but the lowest F1-score, reflecting a precision-recall trade-off. RF exhibited the weakest performance. Confusion matrices indicated significant misclassifications, especially among the Suspicious class.

\subsubsection{Third Experiment: Balanced Dataset with Under-Sampling}
Under-sampling balanced the dataset to 537 tweets equally distributed among classes. Overall performance dropped, with RF surprisingly outperforming others (F1 = 0.375). LR followed closely, while SVM and MNB had slightly lower F1-scores. Confusion matrices revealed increased classification difficulty with balanced but reduced data.

\subsubsection{Fourth Experiment: Balanced Dataset with Over-Sampling}
Over-sampling expanded the dataset to 1,107 tweets equally distributed. All models showed significant improvement, with RF leading (precision 0.814, recall 0.801, F1 0.802), followed by LR and SVM. MNB improved but lagged behind. Confusion matrices showed stronger classification accuracy across all classes.

\subsection{Comparative Analysis with Existing Research}
Comparison with prior studies demonstrates methodological consistency across languages:
Table \ref{tab:ComparissonStudies} Shows the summary comparison of our work with studies with close results of depression detection from social media texts.
\begin{table}[htbp]
    \centering
    \caption{Summary comparison of our work  with close result studies}
    \resizebox{\textwidth}{!}{%
        \begin{tabular}{|l|p{2.8cm}|p{2.5cm}|p{2.2cm}|p{2.2cm}|p{3cm}|}
            \hline
            \rowcolor{lightgray}
            \textbf{\#} & \textbf{Research} & \textbf{Language} & \textbf{Platform(s)} & \textbf{Data} & \textbf{Best Method \& Performance} \\
            \hline
            
            1 & \cite{KadkhodaLR} & English & Twitter & N/A & RF, NB \newline Accuracy = 86\% \\
            \hline
           
            2 & \cite{musleh2022depressionLR} & Arabic & Twitter & 4542 Tweets & RF \newline Accuracy = 82\% \\
            \hline

            3 & This study (2025) & Sorani & Twitter & 1107 Tweets & RF \newline Accuracy = 80\% \\
            \hline
            
        \end{tabular}%
    }
    \label{tab:ComparissonStudies}
\end{table}

While deep learning approaches like LSTM and RNN \cite{DeepLearning} achieve higher accuracy 99\% with large datasets, this study validates that traditional ML models, particularly RF, can achieve competitive results on small, low-resource language datasets, highlighting potential for future work with larger data and advanced techniques.

    \par
\begin{figure}[H]
    \centering    \includegraphics[width=\linewidth]{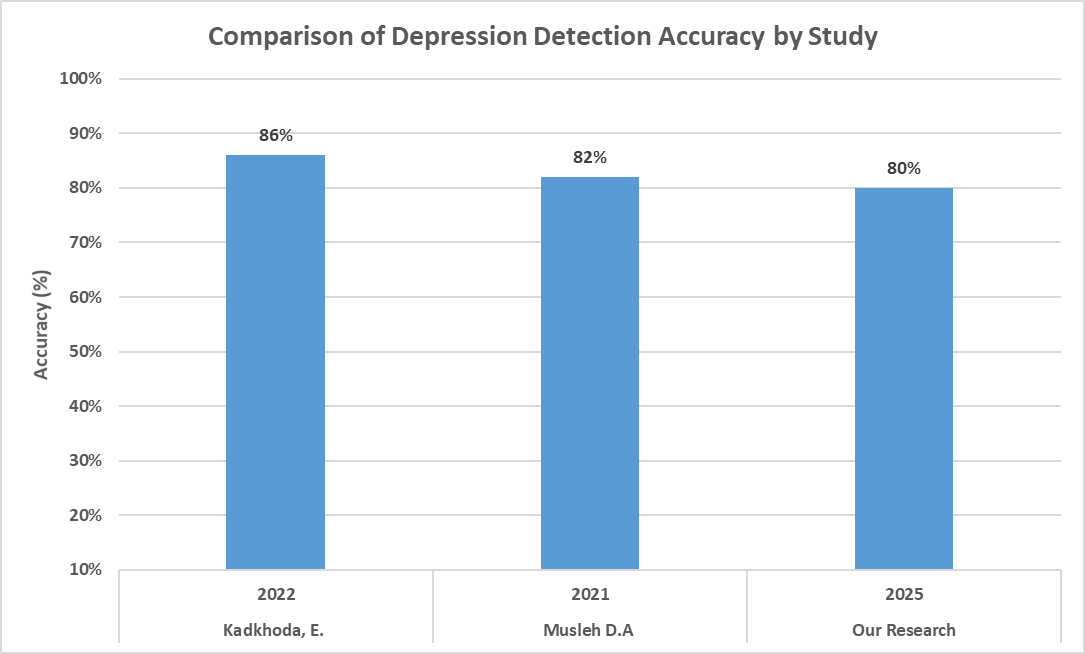}
    \caption{Accuracy comparison of our work with other studies}
    \label{fig:AccuracyComparissionStudies}
\end{figure}

 Figure \ref{fig:AccuracyComparissionStudies} illustrates the accuracy result compression of our research with other studies about depression detection by using ML from social media texts.

\subsection{Discussion}
The four experiments provide a comprehensive comparison of the machine learning models (SVM, MNB, LR, RF) based on accuracy and F1 score, key metrics for classification evaluation. Table \ref{tab:model_scores_integer} shows the accuracy metrics of models for all experiments.

\begin{table}[H]
    \centering
    \caption{Accuracy scores of all models for four experiments}
    \renewcommand{\arraystretch}{1.3}
    \begin{tabular}{|l|c|c|c|c|}
        \hline
        \rowcolor{lightgray}
        \textbf{Model} & \textbf{1st Exp.} & \textbf{2nd Exp.} & \textbf{3rd Exp.} & \textbf{4th Exp.} \\ \hline
        Support Vector Machine      & 67\% & 52\% & 36\% & 77\% \\ 
        Multinomial Naive Bayes     & 66\% & 51\% & 36\% & 70\% \\ 
        Logistic Regression         & 65\% & 50\% & 37\% & 78\% \\ 
        Random Forest               & 66\% & 48\% & 39\% & 80\% \\ \hline
    \end{tabular}
    \label{tab:model_scores_integer}
\end{table}

\par Table \ref{tab:rounded_model_scores} shows the F1-score metrics of models for all experiments.

\begin{table}[H]
    \centering
    \caption{F1 - Scores for each model across four experiments}
    \renewcommand{\arraystretch}{1.2}
    \begin{tabular}{|l|c|c|c|c|}
        \hline
        \rowcolor{lightgray}
        \textbf{Model} & \textbf{1st Exp.} & \textbf{2nd Exp.} & \textbf{3rd Exp.} & \textbf{4th Exp.} \\ \hline
        Support Vector Machine      & 67\% & 48\% & 36\% & 77\% \\ 
        Multinomial Naive Bayes     & 66\% & 45\% & 35\% & 70\% \\ 
        Logistic Regression         & 65\% & 46\% & 37\% & 78\% \\ 
        Random Forest               & 66\% & 44\% & 38\% & 80\% \\ \hline
    \end{tabular}
    \label{tab:rounded_model_scores}
\end{table}

\begin{itemize}
    \item First experiment: All models performed well with SVM leading (accuracy and F1 67\%), indicating balanced precision and recall. MNB and RF were competitive, LR slightly behind. This suggests the dataset was well suited for classification.

    \item Second experiment: Accuracy and F1 dropped across all models due to dataset imbalance and noise. SVM had highest accuracy (51.8\%) but F1 dropped to 47.6\%. RF showed the weakest performance. This reflects challenges posed by the imbalanced dataset.

    \item Third experiment: Performance further declined due to under-sampling, reducing data size, and increasing task complexity. RF outperformed others despite prior weak results, suggesting better robustness to data quality issues. LR remained stable; SVM and MNB struggled.

    \item Fourth experiment: Significant improvements with over-sampling balanced dataset. RF excelled with highest accuracy (80.1\%) and F1 (80.2\%), followed closely by LR and SVM. MNB improved but remained behind. This highlights RF’s ability to model complex patterns when data quality and balance improve.
\end{itemize}

Overall, the experiments revealed that Random Forest emerged as the best performing model, particularly in the final experiment with oversampling, while Logistic Regression demonstrated the most consistent performance across different setups. In contrast, models such as MNB and SVM showed greater sensitivity to data conditions, with their results fluctuating depending on dataset quality and balance. Across all experiments, accuracy and F1 score trends were closely aligned, with the F1 score providing deeper insight into the balance between precision and recall, especially under more challenging data scenarios.

These results underscore the importance of testing models under varied data conditions and highlight that no single model universally outperforms others; performance depends on data quality, feature representation, and task complexity.

\section{Conclusions}
This research aimed to automate the detection of depression in Sorani social media texts using machine learning (ML) and natural language processing (NLP). The first objective was to collect and annotate a comprehensive dataset in Sorani. We collected 960 Twitter posts related to depression and, in collaboration with senior Medicine students studying to receive a degree in Bachelor of Medicine, Bachelor of Surgery (MBBS) at the University of Kurdistan Hewlêr – School of Medicine to annotate the dataset manually.

We developed and evaluated various ML models, including Support Vector Machine (SVM), Multinomial Naive Bayes (MNB), Logistic Regression (LR), and Random Forest (RF), and performed four experimental scenarios. Overall, the results of the models were closely aligned, indicating balanced performance without bias toward any specific class. The oversampling scenario yielded the best results, with RF emerging as the most effective model across most experimental setups. This study is among the first to apply ML and NLP to depression detection in Sorani.

While this research lays the groundwork for automated depression detection in Sorani, several avenues remain open for enhancement, such as dataset expansion to improve model generalizability, exploring deep learning approaches such as BERT, GPT-based models, and convolutional neural networks (CNNs) to potentially improve predictive performance, multimodal analysis by incorporating non-textual data such as images, videos, and emojis alongside text for richer context in depression detection, and extending dialectal coverage to develop more inclusive and robust models for the broader Kurdish-speaking population.

\section*{Data Availability and Companion Website}
The data will be made publicly available at \url{https://kurdishblark.github.io/} in the future.

\section*{Ethical Consideration}
The authors made necessary considerations to protect the privacy of individuals. The annotators provide their consent to publicize their names. The publicly released dataset associated with this study has been processed to remove user handles, aggregate data where appropriate, and remove any information that could be traced back to any particular individual to minimize potential risks to individuals whose tweets were included.

\section*{Author Contributions}
Conceptualization, I.M. and H.H.; methodology, I.M.; software, I.M.; validation, I.M. and H.H.; formal analysis, I.M.; investigation, I.M.; resources, I.M. and H.H.; data curation, I.M.; writing—original draft preparation, I.M.; writing—review and editing, H.H.; visualization, I.M.; supervision, H.H.; project administration, H.H.

\section*{Acknowledgments}
We extend our deep gratitude to Professor Teshk Nouri, Dr. Souzan Hussain, Dr. Naz Ali, Dr. Nasraw Mustafa, and the MBBS (Bachelor of Medicine, Bachelor of Surgery) senior students, studying their final (6th) year at the University of Kurdistan Hewlêr – School of Medicine, for their support in data annotation (for the full list, see the Appendix). Our appreciation is also extended to Dr. Muhammed Qadir for his expert advice on mental health. 

\bibliographystyle{lrec}

\bibliography{SKDepressionDetect}

\begin{thebibliography}{}

\bibitem[\protect\citename{Abdulla and Hama}2015]{SentementAnalysisKurdish}
Abdulla, S. and Hama, M.~H.
\newblock (2015).
\newblock {Sentiment Analyses for Kurdish Social Network Texts using Naive
  Bayes Classifier}.
\newblock {\em Journal of University of Human Development}, 1(4):393--397.

\bibitem[\protect\citename{Ahmadi}2020]{ahmadi-2020-klpt}
Ahmadi, S.
\newblock (2020).
\newblock {KLPT} {--} {K}urdish language processing toolkit.
\newblock In Eunjeong~L. Park, et~al., editors, {\em Proceedings of Second
  Workshop for NLP Open Source Software (NLP-OSS)}, pages 72--84, Online,
  November. Association for Computational Linguistics.

\bibitem[\protect\citename{Alaskar and Ykhlef}2021]{AlaskarLR}
Alaskar, A. and Ykhlef, M.
\newblock (2021).
\newblock {Depression Detection from Arabic Tweets Using Machine Learning
  Techniques}.
\newblock {\em Journal of Computer Science and Software Development}, 2:1--10.

\bibitem[\protect\citename{Aldarwish and Ahmad}2017]{AldarweshLR}
Aldarwish, M. and Ahmad, H.
\newblock (2017).
\newblock {Predicting Depression Levels Using Social Media Posts}.
\newblock {\em In 2017 IEEE 13th international Symposium on Autonomous
  decentralized system (ISADS)}, pages 277--280.

\bibitem[\protect\citename{Almars}2022]{AlmarsLR}
Almars, A.
\newblock (2022).
\newblock {Attention-Based Bi-LSTM Model for Arabic Depression Classification}.
\newblock {\em Computers, Materials \& Continua}, 71(2):3091--3106.

\bibitem[\protect\citename{Amanat \bgroup et al.\egroup }2022]{DeepLearning}
Amanat, A., Rizwan, M., Javed, A.~R., Abdelhaq, M., Alsaqour, R., Pandya, S.,
  and Uddin, M.
\newblock (2022).
\newblock {Deep Learning for Depression Detection from Textual Data}.
\newblock {\em Electronics}, 11(5):676.

\bibitem[\protect\citename{Autumn Asher~BlackDeer \bgroup et al.\egroup
  }2023]{BlackDeer}
Autumn Asher~BlackDeer, MSW, P.~C., David A. Patterson Silver~Wolf, P.,
  Eugene~Maguin, P., and Sara Beeler-Stinn, P.
\newblock (2023).
\newblock Depression and anxiety among college students: Understanding the
  impact on grade average and differences in gender and ethnicity.
\newblock {\em Journal of American College Health}, 71(4):1091--1102.
\newblock PMID: 34242525.

\bibitem[\protect\citename{Badawi \bgroup et al.\egroup }2025]{BadawiLR}
Badawi, S., Kazemi, A., and Rezaie, V.
\newblock (2025).
\newblock {KurdiSent: A corpus for Kurdish sentiment analysis}.
\newblock {\em Language Resources and Evaluation}, 59:601--620.

\bibitem[\protect\citename{De~Choudhury \bgroup et al.\egroup
  }2013]{DeChoudhury}
De~Choudhury, M., Gamon, M., Counts, S., and Horvitz, E.
\newblock (2013).
\newblock {Predicting Depression via Social Media}.
\newblock {\em .}, 7(1):128--137.

\bibitem[\protect\citename{Joshi and Kanoongo}2022]{JOSHI2022217}
Joshi, M.~L. and Kanoongo, N.
\newblock (2022).
\newblock Depression detection using emotional artificial intelligence and
  machine learning: A closer review.
\newblock {\em Materials Today: Proceedings}, 58:217--226.
\newblock International Conference on Artificial Intelligence \& Energy
  Systems.

\bibitem[\protect\citename{Kadkhoda \bgroup et al.\egroup }2022]{KadkhodaLR}
Kadkhoda, E., Khorasani, M., Pourgholamali, F., Kahani, M., and Ardani, A.
\newblock (2022).
\newblock Bipolar disorder detection over social media.
\newblock {\em Informatics in Medicine Unlocked}, 32.

\bibitem[\protect\citename{Lorenzoni \bgroup et al.\egroup
  }2025]{lorenzoni2025assessing}
Lorenzoni, G., Tavares, C., Nascimento, N., Alencar, P., and Cowan, D.
\newblock (2025).
\newblock {Assessing ML classification algorithms and NLP techniques for
  depression detection: An experimental case study}.
\newblock {\em PloS one}, 20(5):e0322299.

\bibitem[\protect\citename{Mahmud \bgroup et al.\egroup }2023]{MahmudLR}
Mahmud, D., Abdalla, B., and Faraj, A.
\newblock (2023).
\newblock {Twitter Sentiment Analysis for Kurdish Language}.
\newblock {\em Qalaai Zanist Journal}, 8(4):1132--1144.

\bibitem[\protect\citename{Mirzaee \bgroup et al.\egroup }2025]{MirzaLR}
Mirzaee, H., Peymanfard, J., Moshtaghin, H., and Zeinali, H.
\newblock (2025).
\newblock {ArmanEmo: a Persian dataset for text-based emotion detection}.
\newblock {\em Language Resources and Evaluation,}, pages 1--23.

\bibitem[\protect\citename{Musleh \bgroup et al.\egroup
  }2022]{musleh2022depressionLR}
Musleh, D., Alkhales, T., Almakki, R., Alnajim, S., Almarshad, S., Alhasaniah,
  R., Aljameel, S., and Almuqhim, A.
\newblock (2022).
\newblock {Twitter Arabic Sentiment Analysis to Detect Depression Using Machine
  Learning}.
\newblock {\em Computers, Materials \& Continua}, 71(2):3463–3477.

\bibitem[\protect\citename{Nadeem \bgroup et al.\egroup }2016]{NadeemDep}
Nadeem, M., Horn, M., and Coppersmith, G.
\newblock (2016).
\newblock {Identifying Depression on Twitter}.
\newblock {\em .}

\bibitem[\protect\citename{Rehmani \bgroup et al.\egroup }2024]{rehmaniLR}
Rehmani, F., Shaheen, Q., Anwar, M., Faheem, M., and Bhatti, S.
\newblock (2024).
\newblock Depression detection with machine learning of structural and
  non‐structural dual languages.
\newblock {\em Healthcare Technology Letters}, 11(4):218--226.

\bibitem[\protect\citename{Roshanaei-Moghaddam \bgroup et al.\egroup
  }2009]{effectsofdepression}
Roshanaei-Moghaddam, B., Katon, W., and Russo, J.
\newblock (2009).
\newblock The longitudinal effects of depression on physical activity.
\newblock {\em General Hospital Psychiatry}, 31(4):306--315.

\bibitem[\protect\citename{Sami and Hassani}2023]{SamiLR}
Sami, M. and Hassani, H.
\newblock (2023).
\newblock {Sentiment Analysis of Opinions about Online Education in the
  Kurdistan Region of Iraq during COVID-19}.
\newblock {\em Qeios}.

\bibitem[\protect\citename{Sinval \bgroup et al.\egroup }2025]{SINVAL2025665}
Sinval, J., Oliveira, P., Novais, F., Almeida, C.~M., and Telles-Correia, D.
\newblock (2025).
\newblock Exploring the impact of depression, anxiety, stress, academic
  engagement, and dropout intention on medical students' academic performance:
  A prospective study.
\newblock {\em Journal of Affective Disorders}, 368:665--673.

\bibitem[\protect\citename{Tejaswini \bgroup et al.\egroup }2024]{Tejaswini}
Tejaswini, V., Sathya~Babu, K., and Sahoo, B.
\newblock (2024).
\newblock Depression detection from social media text analysis using natural
  language processing techniques and hybrid deep learning model.
\newblock {\em ACM Trans. Asian Low-Resour. Lang. Inf. Process.}, 23(1),
  January.

\bibitem[\protect\citename{Tejaswini \bgroup et al.\egroup
  }2025]{ReviewOnDepression}
Tejaswini, V., Sahoo, B., and Babu, K.
\newblock (2025).
\newblock {A Comprehensive Review on Depression Detection Based on Text from
  Social Media Posts}.
\newblock {\em Computational Intelligence for Oncology and Neurological
  Disorders,}, pages 116--132.

\end{thebibliography}

\newpage
\appendix
\setcounter{page}{1}
\renewcommand{\thepage}{A\arabic{page}}
\section*{Appendix - List of MBBS UG6 Students Contributed in Data Annotation}
\setlength{\parindent}{0cm}
\begin{multicols}{2}
\begin{itemize}[itemsep=0.1cm]
\item	Abdulazeez Omar Ali
\item	Abdulmomen M. Abdulhakeem
\item	Ali Kokoi Mahmood 
\item	Ali Nizar Ali
\item	Aram Masood Ibrahim
\item	Arwa Said Mustafa
\item	Bahnam Munthir Rufail
\item	Ban Abdulqader Najeeb
\item	Baran Sardar Pirkhider
\item	Bawan Omer Ibrahim 
\item	Blnd Shukur Saber
\item	Danea Najat Ahmed
\item	Daniya Najat Ali
\item	Diler Talhad Tayib
\item	Dina Dira Broosh
\item	Diyar Abdalkhaliq Qadir
\item	Dwin Qasim Ali
\item	Emam Farouq Emam
\item	Hemin Othman Yaseen
\item	Hezhwan Mohammed Hussein
\item	Honya Halgurd Muhammed
\item	Israa Idres Hamad
\item	Issra Sorani
\item	Jumana Abdulkarim Sami
\item	Khitam Mohsin Mohammedali
\item	Lala Heman Ibrahim
\item	Lana Mokhles Saleh
\item	Lara Kawa Sulaiman
\item	Mahmood Jawdat Jafaar
\item	Mastan Mutasam Fatih
\item	Matelda Majid Mansour
\item	Mina Imran Mohammed Ali
\item	Mina Ramzi Othman
\item	Mohammed Firas Jabbar
\item	Mohammed Osama Ahmed
\item	Mohammed Sabah Abdulrahman
\item	Mohammed Sharef Saleh
\item	Mstafa Diary Rafik
\item	Mustefa Mhemed Adhem
\item	Nina Mariwan Karim
\item	Prusha Dler Abdulrahman
\item	Raghad Dler Hanna
\item	Rameen Isam Anas
\item	Rasar Rebwar Rahim
\item	Razaw Dler Mawlood
\item	Rebaz Othman Wahid
\item	Rooz Mahdi Saleh
\item	Safaa Abdulrahman Ramadhan
\item	Sahand Hussein Abdullah
\item	Sana Sadiq Aziz
\item	Sham Sherko Saeed
\item	Sina Najat Ali
\item	Sitaf Ahmed Ali
\item	Stephany Sammy Benyameen
\item	Talan Abdulla Abdulwahid
\item	Tara Irfan Ali
\item	Tazhna Nawshirwan H. Rafat
\item	Tema Safa Ezzadin
\item	Twana Sulaiman Abdullah
\item	Yad Hatim
\item	Yara Ahmad
\item	Zana Hasan Qader
\end{itemize}
\end{multicols}

\end{document}